\documentclass[conference]{IEEEtran}
\IEEEoverridecommandlockouts
\usepackage{cite}
\usepackage{amsmath,amssymb,amsfonts}
\usepackage{algorithmic}
\usepackage{graphicx}
\usepackage{textcomp}
\usepackage{xcolor}
\usepackage{balance}

\usepackage{todonotes}
\def\BibTeX{{\rm B\kern-.05em{\sc i\kern-.025em b}\kern-.08em
    T\kern-.1667em\lower.7ex\hbox{E}\kern-.125emX}}
\begin{document}

\title{Beyond One-Way Pruning: Bidirectional Pruning-Regrowth for Extreme Accuracy-Sparsity Tradeoff}
\author{\IEEEauthorblockN{Junchen Liu}
\IEEEauthorblockA{
\textit{University of South Florida}\\
Tampa, U.S.A \\
junchen@usf.edu}
\and
\IEEEauthorblockN{Yi Sheng}
\IEEEauthorblockA{
\textit{University of South Florida}\\
Tampa, U.S.A \\
sheng1@usf.edu}
}

\maketitle

\begin{abstract}

As a widely adopted model compression technique, model pruning has demonstrated strong effectiveness across various architectures. However, we observe that when sparsity exceeds a certain threshold, both iterative and one-shot pruning methods lead to a steep decline in model performance. This rapid degradation limits the achievable compression ratio and prevents models from meeting the stringent size constraints required by certain hardware platforms, rendering them inoperable.
To overcome this limitation, we propose a bidirectional pruning–regrowth strategy. Starting from an extremely compressed network that satisfies hardware constraints, the method selectively regenerates critical connections to recover lost performance, effectively mitigating the sharp accuracy drop commonly observed under high sparsity conditions.

\begin{figure*}[htbp]
\centering
\includegraphics[width=\textwidth]{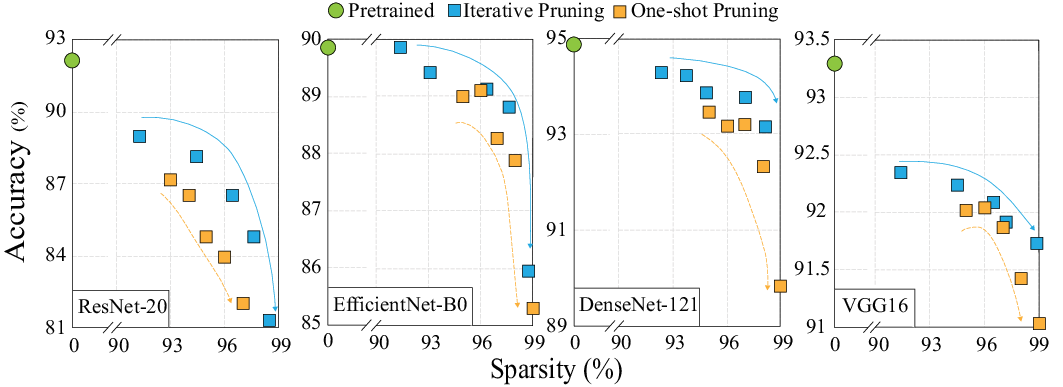}
\caption{Accuracy vs. sparsity on pruning neural networks with iterative and one-shot approach.}
\label{mot1} 
\end{figure*}
\end{abstract}


\section{Introduction}

As artificial intelligence (AI) continues to advance, deep learning models (DNNs) are increasingly deployed on a wide range of edge devices to enable diverse intelligent functionalities \cite{liu2024mobilellm, yao2023mobile}. Nevertheless, deploying such models on resource-constrained devices introduces substantial challenges, particularly in terms of model size \cite{lu2025bluelm}, energy efficiency \cite{han2015learning}, and inference latency \cite{hadji2025edge}. To address these limitations, model compression techniques such as pruning have been extensively explored \cite{lee2020layer, cai2019once}. However, when aiming for extreme compression ratios, model performance typically deteriorates sharply. Consequently, achieving a balance between compactness and performance remains a critical challenge, especially when deploying models on highly resource-limited edge devices.

DNNs have evolved significantly in size over time (e.g., VGG16~\cite{simonyan2014very} and DenseNet-121~\cite{huang2017densely} have tens of millions of trainable parameters), which makes them challenging to deploy on edge devices like smartphones, Raspberry Pi~\cite{raspberrypi}, and Nvidia Jetson Nano~\cite{nvidia_jetson_nano} efficiently. Even tiny models (e.g., ResNet-20~\cite{he2016deep} only holds 270k trainable parameters) aiming at the capability of deployment on edge platforms encountered severe performance degradation during the pruning process. As illustrated in Fig.~\ref{mot1}, accuracy evaluation across ResNet-20, EfficientNet-B0~\cite{tan2019efficientnet}, DenseNet-121, and VGG16 is presented. In each sub-plot, a green-filled, black-outlined circle marker represents the pretrained model. Meanwhile, the blue and orange square markers denote iterative and one-shot pruning processes, respectively. The x-axis indicates the sparsity level (i.e., the zero-value weight ratio), and the y-axis reveals the accuracy of prediction performance on CIFAR-10~\cite{krizhevsky2009learning}. We can observe a similar performance degradation pattern for all listed models. Even for larger networks like DenseNet, 99\% sparsity induces a sharp accuracy drop from 94.86 to 89.83. While these models all achieve a really impressive performance under smaller sparsity (e.g., EfficientNet-B0 trades 0.69 accuracy degradation with 96\% sparsity; DenseNet obtains 93.47 accuracy at 95\% sparsity).

In order to handle the degradation issue at high sparsity, several approaches have been proposed. Pruning On-the-fly~\cite{liu2022pruning} chooses to dynamically adjust pruning masks during training. HRank~\cite{lin2020hrank} proposes to identify and preserve filters that produce high-rank feature maps. ResRep~\cite{ding2021resrep} adopts a structure-aware training scheme that separates the learning of important and redundant filters.

While these solutions have proved their effectiveness, they share several limitations. The criterion HRank utilized requires computing feature maps over batches, which adds considerable overhead for large DNNs. What's more, some layers may inherently be "low-rank” but are more critical. The technique proposed by Pruning On-the-fly might force some parts of the network to become permanently zeroed, losing flexibility to recover. Dependence on well‑chosen hyperparameters poses challenges for ResRep, which requires exhaustive exploration.

\begin{figure}[htbp]
\centering
\includegraphics[width=\columnwidth]{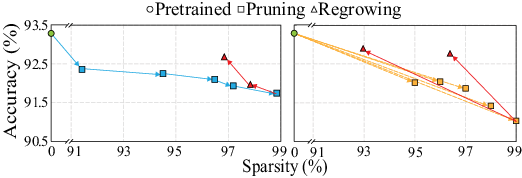}
\caption{Accuracy vs. sparsity for bidirectional pruning-regrowth trajectory on VGG16: (a) iterative approach; (b) one-shot approach.}
\label{mot3} 
\end{figure}
Although modern one-shot and iterative pruning methods employ sophisticated designs to remove uninformative parameters at various granularities, we maintain that this process inevitably eliminates critical parameters. This motivates us to regrow the model starting from a highly sparse state. We evaluated our solution on the VGG16 model, and regrew for both one-shot and iterative strategies. As Fig.~\ref{mot3} shows, blue and orange square markers represent the pruned models obtained by one-shot and iterative pruning, respectively. Moreover, red triangles denote the regrowth results. Both sub-plots reveal the unavoidable performance degradation under extreme compression ratios (i.e., the accuracy degrades from 93.39  (pretrained) to 91.73 by iterative pruning under 98.85\% sparsity and 91.03 by one-shot pruning under 99\% sparsity). With the help of proposed model regrowth, we are delighted to find that degradation can be alleviated. In Fig.~\ref{mot3}(a), starting iterative regrowing the VGG16 model from 98.85\% sparsity, we can recover the prediction performance to 92.64 (96.85\% sparsity). Similarly, we can obtain an impressive performance elevation for one-shot approach, which is revealed in Fig.~\ref{mot3}(b) that we can recover the accuracy to 92.77 (96.39\% sparsity) and 92.9 (92.96\% sparsity). Even at 96.39\% sparsity, our method outperforms the one-shot pruning baseline at 95\%, revealing the sub-optimal weight selection during pruning.

\balance
In this work, we mitigate the performance degradation, which is caused by aggressive pruning, by introducing a model regrowth strategy that selectively restores pruned weights. Beyond enhancing pruning outcomes, we hope our approach contributes to making DNNs more practical for deployment on resource-constrained edge devices. Furthermore, we envision extending this framework to large-scale models such as large language models (LLMs), where maintaining performance while reducing computational demands is increasingly crucial.

\bibliographystyle{IEEEtran}
\bibliography{main}

\begin{thebibliography}{10}
\providecommand{\url}[1]{#1}
\csname url@samestyle\endcsname
\providecommand{\newblock}{\relax}
\providecommand{\bibinfo}[2]{#2}
\providecommand{\BIBentrySTDinterwordspacing}{\spaceskip=0pt\relax}
\providecommand{\BIBentryALTinterwordstretchfactor}{4}
\providecommand{\BIBentryALTinterwordspacing}{\spaceskip=\fontdimen2\font plus
\BIBentryALTinterwordstretchfactor\fontdimen3\font minus \fontdimen4\font\relax}
\providecommand{\BIBforeignlanguage}[2]{{%
\expandafter\ifx\csname l@#1\endcsname\relax
\typeout{** WARNING: IEEEtran.bst: No hyphenation pattern has been}%
\typeout{** loaded for the language `#1'. Using the pattern for}%
\typeout{** the default language instead.}%
\else
\language=\csname l@#1\endcsname
\fi
#2}}
\providecommand{\BIBdecl}{\relax}
\BIBdecl

\bibitem{liu2024mobilellm}
Z.~Liu, C.~Zhao, F.~Iandola, C.~Lai, Y.~Tian, I.~Fedorov, Y.~Xiong, E.~Chang, Y.~Shi, R.~Krishnamoorthi \emph{et~al.}, ``Mobilellm: Optimizing sub-billion parameter language models for on-device use cases,'' in \emph{Forty-first International Conference on Machine Learning}, 2024.

\bibitem{yao2023mobile}
Z.~Yao, J.~Wang, H.~Wu, J.~Wang, and M.~Long, ``Mobile attention: mobile-friendly linear-attention for vision transformers,'' in \emph{Forty-first International Conference on Machine Learning}, 2023.

\bibitem{lu2025bluelm}
X.~Lu, Y.~Chen, C.~Chen, H.~Tan, B.~Chen, Y.~Xie, R.~Hu, G.~Tan, R.~Wu, Y.~Hu \emph{et~al.}, ``Bluelm-v-3b: Algorithm and system co-design for multimodal large language models on mobile devices,'' in \emph{Proceedings of the Computer Vision and Pattern Recognition Conference}, 2025, pp. 4145--4155.

\bibitem{han2015learning}
S.~Han, J.~Pool, J.~Tran, and W.~Dally, ``Learning both weights and connections for efficient neural network,'' \emph{Advances in neural information processing systems}, vol.~28, 2015.

\bibitem{hadji2025edge}
I.~Hadji, M.~Noroozi, V.~Escorcia, A.~Zaganidis, B.~Martinez, and G.~Tzimiropoulos, ``Edge-sd-sr: Low latency and parameter efficient on-device super-resolution with stable diffusion via bidirectional conditioning,'' in \emph{Proceedings of the Computer Vision and Pattern Recognition Conference}, 2025, pp. 12\,789--12\,798.

\bibitem{lee2020layer}
J.~Lee, S.~Park, S.~Mo, S.~Ahn, and J.~Shin, ``Layer-adaptive sparsity for the magnitude-based pruning,'' \emph{arXiv preprint arXiv:2010.07611}, 2020.

\bibitem{cai2019once}
H.~Cai, C.~Gan, T.~Wang, Z.~Zhang, and S.~Han, ``Once-for-all: Train one network and specialize it for efficient deployment,'' \emph{arXiv preprint arXiv:1908.09791}, 2019.

\bibitem{simonyan2014very}
K.~Simonyan and A.~Zisserman, ``Very deep convolutional networks for large-scale image recognition,'' \emph{arXiv preprint arXiv:1409.1556}, 2014.

\bibitem{huang2017densely}
G.~Huang, Z.~Liu, L.~Van Der~Maaten, and K.~Q. Weinberger, ``Densely connected convolutional networks,'' in \emph{Proceedings of the IEEE conference on computer vision and pattern recognition}, 2017, pp. 4700--4708.

\bibitem{raspberrypi}
{Raspberry Pi Foundation}, ``{Raspberry Pi},'' \url{https://www.raspberrypi.com/}, accessed: 2025-10-9.

\bibitem{nvidia_jetson_nano}
{NVIDIA Corporation}, ``{Jetson Nano Developer Kit},'' \url{https://developer.nvidia.com/embedded/jetson-nano-developer-kit}, accessed: 2025-10-09.

\bibitem{he2016deep}
K.~He, X.~Zhang, S.~Ren, and J.~Sun, ``Deep residual learning for image recognition,'' in \emph{Proceedings of the IEEE conference on computer vision and pattern recognition}, 2016, pp. 770--778.

\bibitem{tan2019efficientnet}
M.~Tan and Q.~Le, ``Efficientnet: Rethinking model scaling for convolutional neural networks,'' in \emph{International conference on machine learning}.\hskip 1em plus 0.5em minus 0.4em\relax PMLR, 2019, pp. 6105--6114.

\bibitem{krizhevsky2009learning}
A.~Krizhevsky, G.~Hinton \emph{et~al.}, ``Learning multiple layers of features from tiny images,'' 2009.

\bibitem{liu2022pruning}
D.~Liu and X.~Liu, ``Pruning on-the-fly: A recoverable pruning method without fine-tuning,'' \emph{arXiv preprint arXiv:2212.12651}, 2022.

\bibitem{lin2020hrank}
M.~Lin, R.~Ji, Y.~Wang, Y.~Zhang, B.~Zhang, Y.~Tian, and L.~Shao, ``Hrank: Filter pruning using high-rank feature map,'' in \emph{Proceedings of the IEEE/CVF conference on computer vision and pattern recognition}, 2020, pp. 1529--1538.

\bibitem{ding2021resrep}
X.~Ding, T.~Hao, J.~Tan, J.~Liu, J.~Han, Y.~Guo, and G.~Ding, ``Resrep: Lossless cnn pruning via decoupling remembering and forgetting,'' in \emph{Proceedings of the IEEE/CVF international conference on computer vision}, 2021, pp. 4510--4520.

\end{thebibliography}
\vspace{12pt}
\end{document}